\documentclass[10pt,letterpaper]{article}
\usepackage{hyperref}
\usepackage{cogsci}

\usepackage{subcaption}

\cogscifinalcopy 

\usepackage{pslatex}
\usepackage{apacite}
\usepackage{float} 

\usepackage{makecell}
\usepackage{graphicx}

\title{Dissecting the Ullman Variations with a SCALPEL: Why do LLMs fail at Trivial Alterations to the False Belief Task?}
 
\author{{\large \bf Zhiqiang Pi (owenpi@u.northwestern.edu)} \\
  School of Education and Social Policy, \\ Northwestern University\\ 
  \And {\large \bf Annapurna Vadaparty (avadaparty@ucsd.edu)} \\
  Department of Cognitive Science, \\ University of California, San Diego\\
  \AND {\large \bf Benjamin K. Bergen (bkbergen@ucsd.edu)} \\
  Department of Cognitive Science, \\ University of California, San Diego\\
  \And {\large \bf Cameron R. Jones (cameron@ucsd.edu)} \\
  Department of Cognitive Science, \\ University of California, San Diego\\
}

\begin{document}

\maketitle

\begin{abstract}
Recent empirical results have sparked a debate about whether or not Large Language Models (LLMs) are capable of Theory of Mind (ToM). While some have found LLMs to be successful on ToM evaluations such as the False Belief task, others have shown that their performance is not robust against trivial alterations to stimuli.
In this paper, we introduce SCALPEL---a technique to  incrementally modify stimuli to test different specific hypotheses about why LLMs fail---and apply this method to the `transparent-access' modification of the unexpected contents task. 
Our results suggest that LLMs often do poorly because they fail to make essential common-sense inferences, such as that seeing a transparent container implies recognizing its contents. We conclude that while modern LLMs go beyond mere pattern matching, they still fall short of robust human-like ToM.
We argue that SCALPEL can help cognitive scientists examine LLMs' capabilities in finer detail and provide insight into alternative mechanisms by which tasks that are used to assess human cognition might be completed.

\end{abstract}

\section{Introduction}

Due to the superficially human-like behavior of LLMs, researchers are increasingly repurposing tasks designed by psychologists to measure human cognitive abilities and administering them to LLMs. This approach, sometimes referred to as machine psychology \cite{hagendorff2023machine}, uses established instruments that not only enable AI researchers to explore emergent capabilities of LLMs, but also have the potential to provide cognitive scientists with novel insights into human cognition \cite{binz2023turninglargelanguagemodels, centaur2024}. Finegrained analysis of successes and failures of these models can also guide further development in machine learning and inform applications of machine learning models in the real world. In this paper, we introduce Selective Comparison of Adversarial Linguistic Prompts to Explain Lacunae (SCALPEL): a technique to understand specifically where and why LLMs fail at machine psychology tasks.

In the present work, we focus on machine psychology tasks designed to examine Theory of Mind (ToM): the ability to reason about the unobservable mental states of other agents \cite{Premack_Woodruff_1978}. Various studies aiming to evaluate the ToM capabilities of LLMs have produced inconsistent conclusions
\cite{kosinski2024evaluating, ullman2023large, shapira2023clever, kim2023fantom, gandhi2023understanding, jones2023epitome, strachan2024testing}. 
To gain insight into this inconsistency, we apply SCALPEL on the Transparent-Access alteration of the Unexpected Contents Task. The Unexpected Contents Task is a commonly used test of children's ToM development \cite{unexpectedcontents}. Typically, a child is shown a container with a label that is inconsistent with its contents. Then the child is asked what another child who has no prior knowledge of the container will believe its contents are. To answer this correctly, the child must be able to realize that the other child doesn't know that the label is inconsistent. \citeA{kosinski2024evaluating} adapted this task to evaluate ToM capabilities of LLMs; an example prompt from their study is below:

\begin{quote}
\begin{verbatim}
In the freezer, there is a container 
filled with ice cream. There is no jam 
in it. Yet, the label says "jam" and 
not "ice cream". The label is wrong. 
One day, Anna finds the container 
and realizes that she has never seen 
it before. She reads the label. She 
is delighted to have found this 
container.

Question: Fill in the blank with 
the best option. She loves eating ___ 
- ice cream
- jam

Answer:
\end{verbatim}
\end{quote}

Since the label of the container says ``jam'' and Anna has no other means to know what the true contents of the container are, one could reasonably infer that Anna believes that the container contains jam. Moreover, because she is delighted to have found this container, she must love eating jam. \citeA{kosinski2024evaluating} reported that GPT-4 was able to solve this task 90\% of the time, while independent researchers had found similar successes for LLMs on other false-belief tasks \cite{trott2023}.

As \citeA{kosinski2024evaluating} suggests that ToM capabilities may have emerged in LLMs, \citeA{ullman2023large} set out to examine the robustness of LLM performance using a number of modified versions of the task, including a `Transparent-Access Variation' in which the container is explicitly described as transparent. (See \textbf{original} in Table ~\ref{tab:modifications}.) With this modification, Anna can now see the true contents of the container, so it can be inferred that Anna is delighted to have found the container because she loves eating \textit{ice-cream}, the true contents of the container. Ullman reported that GPT-3.5 incorrectly assigned $95\%$ probability to the label being the completion of the item. Inspired by this finding, \citeA{shapira2023clever} performed a systematic evaluation of vignettes used by \citeA{kosinski2024evaluating} with adversarial modifications suggested by \citeA{ullman2023large}, and they found that both GPT-3.5 and GPT-4 were correct only $18.8\%$ of the time on the transparent access modification of the unexpected contents task.

Failures with seemingly trivial modifications like this one have led to the interpretation that LLMs rely on shallow heuristics and spurious correlations rather than genuine ToM capabilities to solve false belief tasks \cite{ullman2023large, shapira2023clever}. 
On this account, LLMs are only able to provide successful responses to False Belief questions because they bear a strong superficial similarity to examples that appear in their training set. However, \citeA{hu2024auxiliarytaskdemandsmask} suggest that this degradation in performance might be attributable to the increased task demands caused by adversarial modifications \cite{hu_re-evaluating_2025}. Following this argument, an alternative possibility is that LLMs’ failure may be the result of the inability to make other common inferences that are required to complete the task \cite{bloomTwoReasonsAbandon2000}, such as the inference that a transparent container affords seeing its contents. 

To adjudicate between these hypotheses, we apply SCALPEL to make minor incremental modifications to the Transparent-Access Variation of the Unexpected Contents Task. Each modification represents a different hypothesis about why LLMs might be failing. We test each hypothesis by removing potential sources of failure. For instance, to measure the extent to which models fail because their responses are not sensitive to the fact that a transparent container allows observers to see its contents, we make this inference explicit in the text. To the extent that the modification improves model performance, it suggests that this bridging inference was a point of failure for the model.
We use this technique to measure the contribution of different sources of error (from physical inferences like the one above, to more psychological inferences, like that looking at a transparent container implies recognizing its contents). We show that SCALPEL can help explain behavior by shedding light on the component operations performed by LLMs when solving cognitive tasks \cite{hu_re-evaluating_2025}.

\section{Method}

All analyses were preregistered and all materials are available online.\footnote{\url{https://osf.io/td3fw/}} \footnote{\url{https://github.com/UCSD-Language-and-Cognition-Lab/scalpel_transparent}}

\subsection{Materials}

\begin{table*}
\begin{center} 
\vskip 0.12in
\begin{tabular}{clclr} 
\hline
Modification   &  Stimulus  &  GPT3.5 & GPT4 \\
\hline
original     &  \makecell[l]{...there is a \textbf{transparent} container filled with ice cream...}  &  22.14\% & 20.36\% \\
\hline
see-through     &  \makecell[l]{...there is a \textbf{see-through} container filled with ice cream...}  &  18.57\% & 20\% \\
\hline
see-inside     &  \makecell[l]{...there is a transparent container filled with ice cream \textbf{that anyone} \\ \textbf{can see inside of}...}  &  18.92\% & 20.36\%\\
\hline
read\_look     &  \makecell[l]{...She reads the label. \textbf{Then, she looks at the container}...}  &  37.14\% & 40.36\% \\
\hline
look\_read     &  \makecell[l]{...She \textbf{looks carefully at the container and then} reads the label...}  &  32.86\% & 36.07\% \\
\hline
recognize\_content     &  \makecell[l]{...She reads the label. \textbf{Then, she looks at the container} \\ \textbf{and recognizes what is inside}...}  &  54.28\% & 89.64\% \\
\hline
recognize\_label & \makecell[l] {...She reads the label. Then, she looks at the container \\ and recognizes what \textbf{it says}...}  &   & 27.14\% \\
\hline
visualize & \makecell[l] {...She reads the label. Then, she looks at the container \\ and \textbf{visualizes} what is inside...}  &   & 55.71\% \\
\hline
\end{tabular} 
\caption{Modifications to the Unexpected Contents Task and corresponding accuracy of GPT3.5 and GPT4.}
\label{tab:modifications} 
\end{center} 
\end{table*}

Our proposed method, SCALPEL, involves generating hypotheses and counter-hypotheses about the inferences that the LLMs might be failing to implicitly make, creating minimally invasive modifications to the original prompts based on these hypotheses, and analyzing the different levels of performance exhibited by LLMs on the modified prompts. This technique adapts a powerful paradigm from psycholinguistics---the use of tightly-controlled minimal pairs \cite{frazier1982making}---to the new challenge of probing large language models. Psycholinguistic techniques have already proven valuable in identifying model capabilities \cite{marvin2018targeted} and in generating adversarial examples \cite{naik2018stress, mccoy2019right}. Here we extend this tradition by generating minimal pairs to test a range of specific hypotheses about when and why models fail.

The hypotheses and modifications made were as follows:

\textbf{\textit{Transparent implies Visible Contents}} LLMs might fail to adjust their answers when given the Transparent Modification of the task because they don’t implicitly infer that people are able to see through transparent containers. To test this hypothesis, we make the following modifications. 1) We exchange the word “transparent” for the more explicit “see-through” (see \textbf{see-through} in Table ~\ref{tab:modifications}). 2) We make the meaning of “transparent” even more explicit by adding the clause ``that anyone can see inside of'' (\textbf{see-inside}, Table ~\ref{tab:modifications}).

\textbf{\textit{Reading the Label Implies Looking at the Container}} While LLMs might properly represent ``transparent'', they might not be sensitive to the inference that when reading the label of a transparent container, the character also sees its contents. To examine this possibility, we append an explicit statement that the character looks at the container after reading the label (\textbf{read\_look}, Table ~\ref{tab:modifications}). A possible explanation for a positive effect of this modification is that the object inside the container is made more salient than the label as it is more recently mentioned \cite{gernsbacher2013language}. To test this, we introduce another variation stating that the character looks inside of the container before stating that they read the label (\textbf{look\_read}, Table ~\ref{tab:modifications}).

\textbf{\textit{Seeing implies Recognizing}} Even if an LLM is appropriately sensitive to the inference that reading the label on a transparent container will lead to the character looking inside of the container, it may not be sensitive to the inference that the character was able to recognize what they see. To address this possibility, we add another sentence in the stimuli to explicitly state that the character in the story recognizes what is inside the container (see \textbf{recognize\_content}, Table ~\ref{tab:modifications}). 

\subsection{Procedure}

Our experimental procedure mostly aligns with the procedure outlined in \citeA{shapira2023clever}.
We probed LLMs in a zero-shot fashion with the prompts used in the the Transparent-Access condition of their ADVersarial CommonSense with False Belief dataset, which also formed the basis of our modifications. Each scenario followed a preprompt of an unrelated question with a similar format, and was followed by a fill-in-the-blank question. We used the OpenAI API to elicit a response of no more than 30 tokens from the models. 

We diverged from the procedure used in \citeA{shapira2023clever} to use a temperature of 1 instead of 0. While a temperature of 0 guarantees that the model selects the most likely token, we were interested in a more fine-grained analysis of the distribution of errors that the model might make. We therefore sampled from the models' output distributions multiple times at a temperature of 1 for each item in order to estimate their error rate. A simulated power analysis indicated that 20 samples per model per item would provide sufficient power to detect moderately sized effects. As well as providing more insight into the model's error distribution, this technique also allows us to test the robustness of our modifications.

We counterbalanced the objects referred to as the contents and the label of the containers to control for the possibility that some objects were more associated with some containers. For each probe, we recorded whether the model's response exactly matched the correct answer. We applied this procedure on gpt-3.5-turbo-0301 and gpt-4-0613 with and without the modifications described in the previous section.

\subsection{Statistical Analysis}

To evaluate the impact of each modification on model performance, we fit a mixed effects logistic regression model predicting accuracy on the basis of whether a modification is added (modified vs. original) with random intercepts for each scenario (the original passages from which our items are formed).

\section{Results}

\begin{figure*}[ht]
\begin{center}
\includegraphics[width=0.9\linewidth]{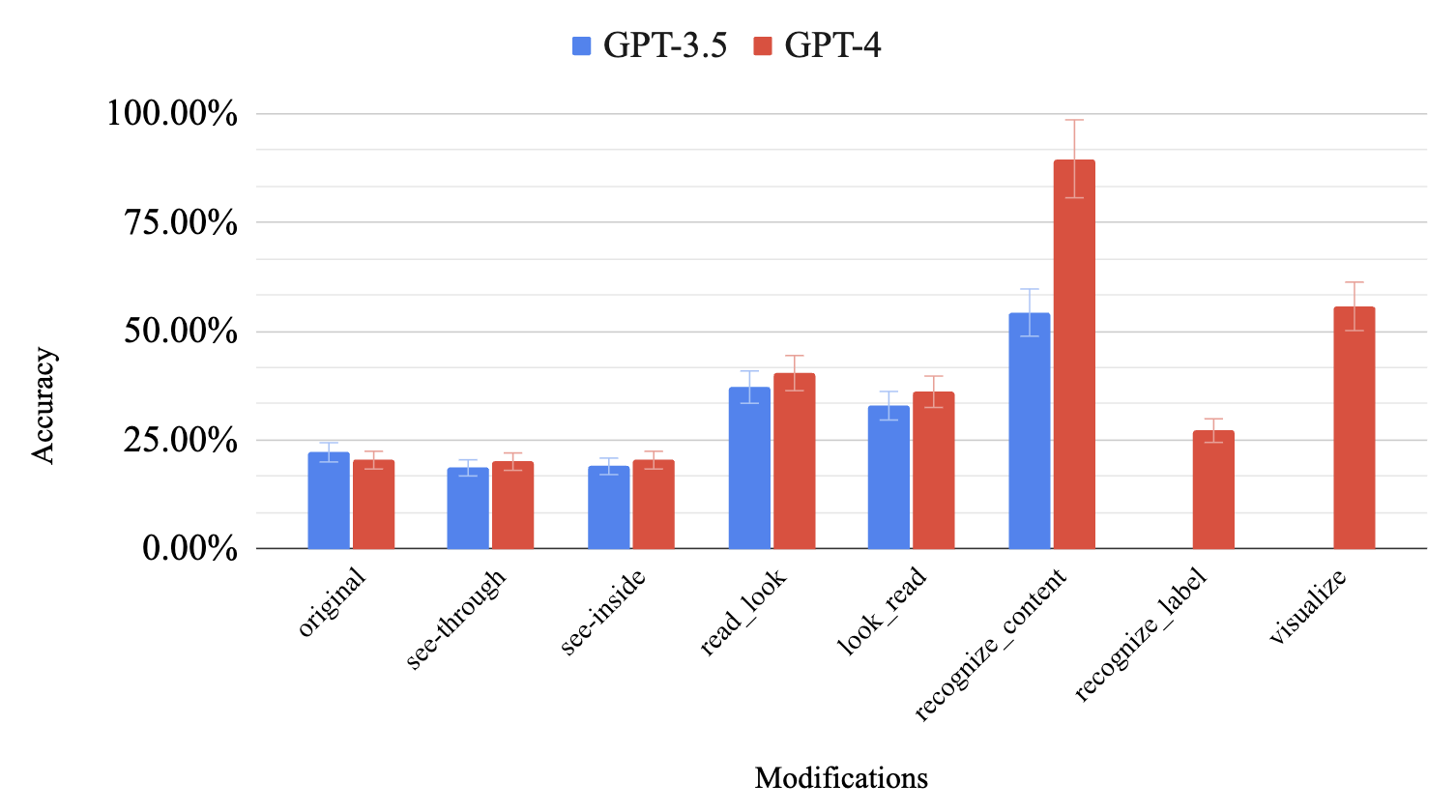}
\end{center}
\caption{Accuracy rates of both GPT-3.5 and GPT-4 on the original Transparent-Access modification of the Unexpected Contents task and additional modifications which included connecting inferences. The high accuracy achieved by GPT-4 on the \textbf{recognize\_content} modification, in addition to the small improvements from the \textbf{read\_look, look\_read, recognize\_label, visualize} modifications, suggest that the model is failing to make the inference that characters recognize the content of the transparent container when they look at it.} 
\label{fig:interface}
\end{figure*}

First, we replicated findings of \citeA{shapira2023clever}; GPT-3.5 and GPT-4 both achieved $\sim20\%$ accuracy on the original Transparent-Access Variation, as compared to 18.8\% for both models as reported in \citeA{shapira2023clever}. 

Second, we found no significant accuracy difference between the original modification and \textbf{see-through} (GPT-3.5: $z=-1.476, p=0.14$, GPT-4: $z=-0.195, p=0.85$) or \textbf{see-inside} (GPT-3.5: $z=-1.312, p=0.19$, GPT-4: $z=0.000, p=0.99$).

Third, explicitly stating that the character looks at the container produced improved accuracy over the original modification in both the \textbf{read\_look} (GPT-3.5: $z=5.825, p<0.001$, GPT-4: $z=9.898, p < 0.001$) and  \textbf{look\_read} (GPT-3.5: $z=4.246, p<0.001$, GPT-4: $z=7.568, p<0.001$) modifications. However, even with these modifications, both LLMs still perform below chance at $\sim 35\%$.

Lastly, the \textbf{recognize\_content} modification significantly improved the performance of both GPT-3.5 ($z=11.282, p<0.001$) and GPT-4 ($z=30.59, p<0.001$). While GPT-4 is able to achieve about $90\%$ accuracy, GPT-3.5 performs only slightly above chance.

\section{Additional Experiments}

The gain in performance produced by the \textbf{recognize\_content} modification could suggest that this is the crucial inference that GPT-4 is failing to make when it fails at the transparent access alteration. However, there are other features of our modification which could provide alternative explanations for its success.
One benefit of the SCALPEL method is that it is easily extensible to iteratively test novel explanations, by designing additional modifications which differ minimally from the critical stimulus in the relevant feature. We formulate our hypotheses and novel modifications below. Note that as these modifications were designed as a follow-up to the positive results above; therefore, these additional experiments were not pre-registered. We followed the same experimental procedure with the same version of GPT-4. However, due to the version of GPT-3.5 used in prior experiments being deprecated, these experiments were not performed for GPT-3.5.

\textbf{\textit{Direct Reference to Mental State}} Theory of Mind involves the modeling of others' mental states using observations of their behavior. As such, the performance improvement seen with the \textbf{recognize\_content} modification may be due to an explicit reference to the unobservable mental state of the character rather than the specific implication of the character recognizing the contents of the container. To test this hypothesis, we created the \textbf{visualize} modification to state that the character visualizes the contents of the container (see Table ~\ref{tab:modifications}). While also containing an explicit reference to the character's mental state, it doesn't contain information relevant to the prompt as the \textbf{recognize\_content} modification does. To the extent that the \textbf{recognize\_content} modification increases performance over the \textbf{visualize} modification, it would suggest that the performance gain is due to clarifying a specific inference---that the character recognizes what they see inside the container---rather than from a general increase in the salience of the character's mental states.

\textbf{\textit{Distance from mention of label}} The \textbf{read\_look}, \textbf{look\_read}, and \textbf{recognize\_content} modifications contain an additional sentence, and so are not controlled for length with the original.
Importantly, this also increases the distance between the last mention of the label and the model's response.
It could be that this increase in distance lowers the likelihood that the model generates the response which is related to the label, creating a confound with our explanation that the explication of inferences plays a role here. In order to address this confound, we created an additional \textbf{recognize\_label} condition (Table ~\ref{tab:modifications}) by changing the \textbf{recognize\_content} modification to state that the character recognized the label of the container rather than its contents. The \textbf{recognize\_content} and \textbf{recognize\_label} modifications  differ by only one word, allowing us to isolate the contribution of making the recognition of the content inference explicit while controlling for length.

\subsection{Results}

Results from these additional modifications demonstrate that these alternative hypotheses are insufficient to explain the performance improvement under the \textbf{recognize\_content} condition for GPT-4. Specifically, GPT-4 performs significantly better on the \textbf{recognize\_label} modification than in the \textbf{original} condition ($z=3.86, p<0.001$), but significantly worse than it does on the \textbf{recognize\_content} modification ($z=-27.082, p<0.001$). Similarly, the \textbf{visualize} modification improves accuracy over the \textbf{original} condition ($z=14.205, p<0.001$), but GPT-4 performs significantly worse on this modification than on the \textbf{recognize\_content} modification ($z=-13.779, p<0.001$).

\begin{figure*}[htbp]
    \centering
    \begin{subfigure}[b]{0.45\textwidth}
         \centering
         \includegraphics[width=\textwidth]{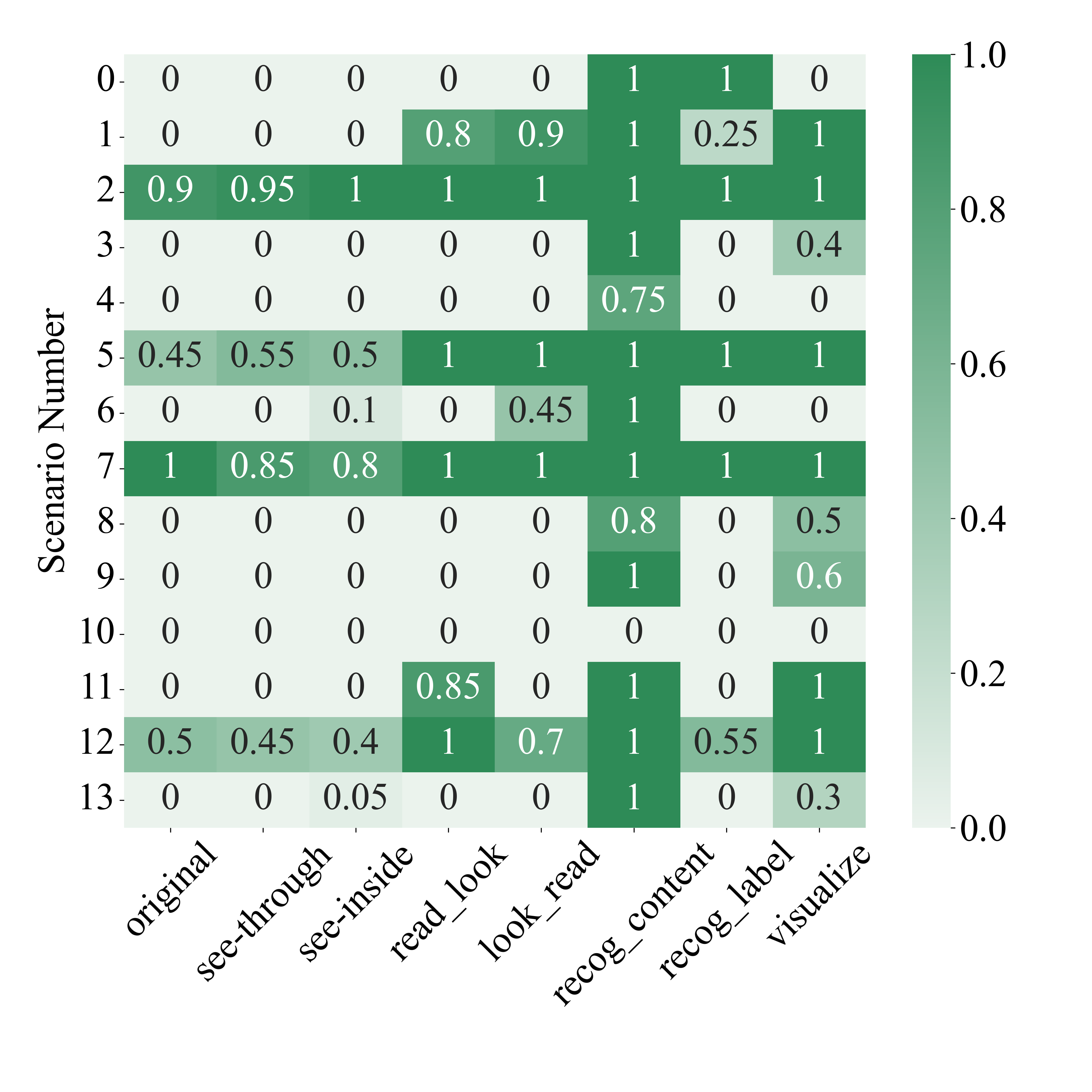}
         \caption{GPT-4 By-Item Accuracy}
         \label{fig:graph1}
    \end{subfigure}
    \hspace{0.02\textwidth}%
    \begin{subfigure}[b]{0.45\textwidth}
         \centering
         \includegraphics[width=\textwidth]{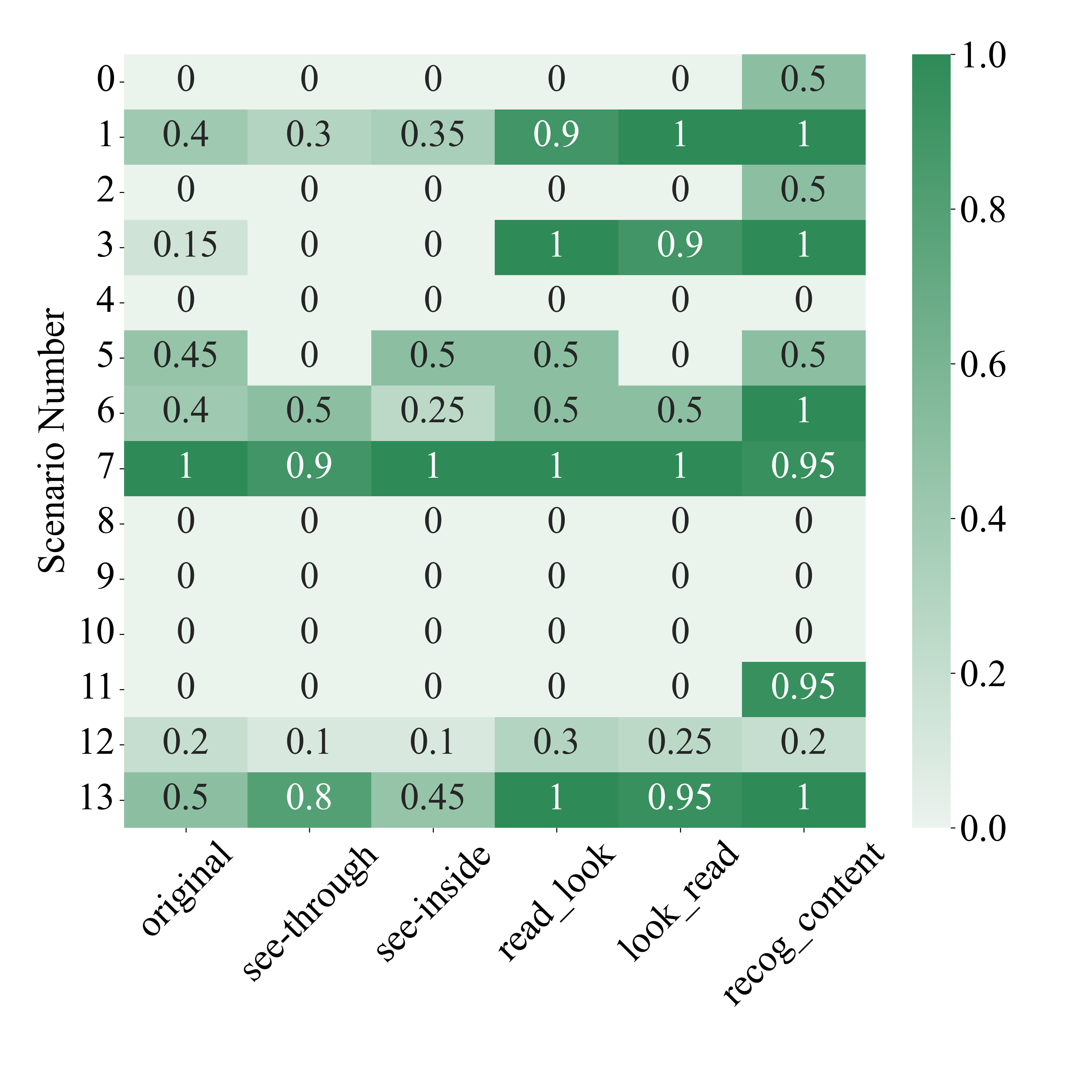}
         \caption{GPT-3.5 By-Item Accuracy}
         \label{fig:graph2}
    \end{subfigure}
    \caption{Each of the scenarios are tested for each model 20 times for each modification. Each cell in the heatmap represents the accuracy of the corresponding model on each item.}
    \label{fig:two_graphs}
\end{figure*}

\section{Discussion}

Using the SCALPEL technique, we identified more specific explanations for LLMs' inconsistent performance on False Belief task variations observed in prior work, and were able to adjudicate among them.
Making it more explicit that the container can be seen through does not improve model performance. Although this does not exclude the possibility that the model doesn't have a sufficient understanding of the word "transparent", this implies that a lack of understanding about transparency is not responsible for model failure.
By contrast, LLMs' performance improved as we made explicit the inference that by reading the label of the container, a person will likely also look at the container, and thus at its contents. 
LLM accuracy significantly increased when the fact of the person looking at the container was mentioned explicitly, suggesting that this inference is a weak point in the process by which LLMs generate their responses.
This effect was robust whether or not looking at the container is mentioned first. However, this modification did not improve model performance above chance. This indicates that even though the modification makes the models less likely to produce the incorrect answer, other inferences required to produce the correct response are still missing.

We see the most drastic improvements in model performance, especially in GPT-4, under the \textbf{recognize\_content} modification. Using a similar measure used by \citeA{kosinski2024evaluating}, GPT-4 in the \textbf{recognize\_content} condition was able to solve $11/14 = 0.79\%$ of the scenarios, approaching the $90\%$ performance reported by \citeA{kosinski2024evaluating} without adversarial modifications. This could indicate that LLMs are failing at the Transparent Access variation of the Unexpected Contents task because their representation of a sentence saying that a person `looks at' a transparent container does not incorporate the likely (to humans) inference that the person can recognize its contents.

Our additional experiments suggest that alternative hypotheses are insufficient to explain the extent of GPT-4's performance improvement on the \textbf{recognize\_content} condition. GPT-4 demonstrates improved performance on the \textbf{recognize\_label} and the \textbf{visualize} modifications over the \textbf{original} modification, indicating that the increased distance from the mention of the label and an explicit reference to the character's mental state contributed to the improvement in performance seen in the \textbf{recognize\_content} condition. However, they only offer partial explanations for GPT-4's improvement with the \textbf{recognize\_content} condition as GPT-4's accuracy on these modifications is significantly worse than its performance on the \textbf{recognize\_content} modification. 

Additionally, our results suggest that this improvement is unlikely to result from brittleness: stochastic changes in output due to any kind of changes in the prompt. As seen in Figure \ref{fig:graph1}, performance improvements are seen in $13/14$ scenarios (including scenario 7 where GPT-4 achieved $100\%$ accuracy in the original  for the \textbf{recognize\_content} condition. This suggests that the improvement from the \textbf{recognize\_content} modification is generalizable across contexts, rather than simply stochastic.

Although this modification improves GPT-4 performance to almost 90\%, it only pushes GPT-3.5 to slightly above chance. The \textbf{recognize\_content} modification significantly improved GPT-3.5 accuracy versus the \textbf{read\_look} condition ($z=6.783, p<0.001$), suggesting that the corresponding inference added meaningful information. However, it is likely that GPT-3.5 is also lacking in other key inferences required to respond with the correct answer. It is possible that there may be important differences in the internal computations employed by the two different models to solve the Unexpected Contents task under different modifications. Future work should explore this potential difference in a wider range of LLMs to understand potential qualitative differences in their internal computation that causes surface level quantitative differences in performance on benchmarks.

As well as addressing specific questions about the features of items which might cause models to fail, our work also addresses a broader question about whether LLMs can \textit{only} solve false belief-like tasks using superficial pattern matching. This was one potential implication of \citeA{ullman2023large}'s study: the fact that LLMs fail at trivial alterations suggest that they do not display robust ToM abilities, they only succeed at false belief tasks that are superficially similar to training items. Our results suggest that it is unlikely that recent LLMs exploit \textit{solely} superficial cues to solve false belief tasks. Our modifications are no more similar to the prompts used by \citeA{kosinski2024evaluating} than those used in \citeA{shapira2023clever}. It is therefore unlikely that LLMs are performing better on our modifications due to their similarity to training examples.

However, our results are also supportive of the idea that any ToM abilities displayed by these models are not robust. Inference such as 
`looking at a transparent container implies recognizing its contents' and (to a lesser extent) `reading a container's label implies seeing the container' are arguably crucial parts of a Theory of Mind. The fact that models performed better when supplied with these inferences suggests that the way the models were encoding these sentences did not intrinsically generate a representation of these perceptual and mentalistic aspects of the scenario. A reasoner with a robust Theory of Mind should be capable of making these inferences. In short, models appear to be doing something more sophisticated than pattern matching, but less robust than human ToM.

Moreover, these result offers support to the argument that adversarial modifications can cause auxiliary task demands which may mask core capabilities being examined \cite{hu2024auxiliarytaskdemandsmask, hu_re-evaluating_2025}. We believe that SCALPEL can be a powerful tool to alleviate auxiliary demands introduced by adversarial modifications and focus evaluations on the core capabilities of interest. The results highlight the value of going beyond assessing LLM accuracy in evaluating their performance.
LLMs continue to show brittle performance across a variety of tasks that human comprehenders solve capably \cite{kim2023fantom, gandhi2023understanding, shapira2023clever, Mitchell_2023}.
Our proposed method, SCALPEL---creating targeted, minimal modifications to error-producing stimuli to understand which aspects of the stimulus pose a challenge for LLMs---can be useful for pinpointing the reason why models succeed or fail on a wide range of psychological tasks to uncover their internal computation \cite{hu_re-evaluating_2025}.

Finally, applying SCALPEL for machine psychology tasks could allow cognitive scientists to gain more general insights into how psychological tasks can be completed with high accuracy.
In some cases, these techniques may help to shed light on differences between humans and machine learning systems--- highlighting instances where models fail for using superficial heuristics.
In other cases, it could help to identify where human participants could also be using heuristic strategies.
\citeA{Strachan2024} reported that human participants were also relatively unsuccessful in solving the Unexpected Contents task with the transparent access modification.
SCALPEL provides an inexpensive way to test different hypotheses about why a system might fail on variations of a stimulus that could be used to motivate research on human language comprehension: targeting specific computations that might underlie intelligent language comprehension behaviour across a diverse systems.

\section{Limitations}
Looking at Figure. \ref{fig:two_graphs}, we observe that there are noticeable by-item differences for both GPT-3.5 and GPT-4. Although not directly explored in this paper, we believe that future work can apply the SCALPEL method similarly to better understand why some prompts appear to be easier for LLMs while others are more difficult. These studies may help create a more standardized approach to studying LLM capabilities. 

We explored only seven of an infinite number of possible variations to the stimuli. Moreover, this leaves open the possibility that the \textbf{recognize\_content} modification improved model performance for reasons beyond our proposed alternative hypotheses. Future work may help explain the failure of GPT-3.5 under the \textbf{recognize\_content} condition and better understand the capabilities of LLMs.

While our results allows for an intuitive explanation for the impact of different modifications, it may not always be possible. Researchers applying SCALPEL to test a wider range of hypotheses and on a larger variety of LLMs may observe patterns of performances that doesn't afford a clear interpretation. Such results can illuminate important differences between human cognition and LLM processing. However, this is not explored in the current work.

\section{Conclusion}

We introduced SCALPEL to pinpoint why LLMs fail on psychological tasks and applied the technique to investigate Theory of Mind. We found that explicitly stating implicit inferences that humans commonly make improves LLMs' performance significantly. This finding calls for additional scrutiny to stimuli developed for psychological tasks to probe LLM capability, and it points to the importance of careful analysis to understand their successes and failures. Failing at a task meant to measure Theory of Mind may not entail the absence of a capacity for Theory of Mind. More detailed examinations of LLMs' response patterns using SCALPEL allow a deeper understanding of the specific extents of emergent capabilities of LLMs and enable cognitive scientists to use LLMs to test alternative explanations for ways in which psychological tasks can be solved.

\section*{Acknowledgments} The authors would like to thank Sean Trott for helpful input at various stages of this project. Also, we appreciate the anonymous reviewers for their insightful comments and suggestions, which greatly improved the quality of this paper.

\bibliographystyle{apacite}

\setlength{\bibleftmargin}{.125in}
\setlength{\bibindent}{-\bibleftmargin}

\bibliography{custom}

\end{document}